# A perceptual bias of AI Logical Argumentation Ability in Writing

Xi Cun[1]*, Jifan Ren[2]*[3], Asha Huang[1], Siyu Li[1], Ruzhen Song[2]

**Abstract:**

**"Can machines think?"** is a central question in artificial intelligence research. However, there is a substantial divergence of views on the answer to this question. Why do people have such significant differences of opinion, even when they are observing the same real-world performance of artificial intelligence? The ability of logical reasoning like humans is often used as a criterion to assess whether a machine can think. This study explores whether human biases influence evaluations of AI's reasoning abilities.

An experiment was conducted where participants assessed two texts on the same topic—one AI-generated and one human-written—to test for perceptual biases in evaluating logical reasoning. Based on the experimental findings, a questionnaire was designed to quantify participants' attitudes toward AI.

The results reveal a bias in perception. People's evaluations of the logical reasoning ability of AI-generated texts are significantly influenced by their preconceived views on the logical reasoning abilities of AI. Furthermore, frequent AI users were less likely to believe that AI usage undermines independent thinking.

This study highlights the need to address perceptual biases to improve public understanding of AI's capabilities and foster better human-AI interactions.

## INTRODUCTIOIN

In his landmark paper, Turing (1) posed the question: **Can machines think?** He proposed the famous Turing Test, which suggests that if a machine can engage in a conversation with a human (via a teletype device) without being identified as a machine, it can be considered intelligent. In 2023, the large language model (LLM) ChatGPT reportedly passed the Turing Test. However, opinions on this achievement are divided. Some argue that its algorithms already demonstrate reasoning and understanding, while others remain cautious. This has sparked widespread interest in exploring the true capabilities of large language models (2).

The focus, however, should not be on whether an LLM resembles human thought but on whether it can exhibit intelligence. If ChatGPT does not think like a human, can we still claim that a machine can think?

Since humans consider the ability to use language for logical reasoning a key indicator of intelligence, numerous tests have been conducted to assess the logical reasoning abilities of large language models (LLMs).

---

[1] School of Humanities and Social Science, Harbin Institute of Technology, Shenzhen, China.
[2] School of Economics and Management, Harbin Institute of Technology, Shenzhen, China.
[3] *Corresponding author. email: cunxi@hit.edu.cn; renjifan@hit.edu.cn.

**Tests on AI's logical reasoning ability.**

The logical reasoning capabilities of artificial intelligence have become a popular yet controversial topic (3-7). In recent years, AI based on LLMs has made significant progress in logical reasoning, but its abilities still have limitations. Studies show that while these models perform excellently on tasks within the scope of their training data, their performance significantly declines when faced with novel or out-of-distribution problems (4,7). Furthermore, research suggests that LLMs do not possess genuine logical reasoning abilities but merely mimic reasoning patterns found in training data (8). ChatGPT excels in simple deductive reasoning but has difficulties with abductive reasoning and tasks requiring semantic understanding (6). A study from Ohio State University revealed that LLMs like ChatGPT often fail to defend correct answers in debate-style dialogues, accepting invalid arguments instead (9). The "Chain of Thought" prompting technique can significantly improve AI's reasoning accuracy, with improvements up to 100% (10), but does not bridge the gap to human cognition. The capabilities and limitations of artificial intelligence in logical reasoning remain highly debated. Even technical experts have doubts about the capabilities and limitations of artificial intelligence in logical reasoning. Meanwhile, the overhyped marketing of AI's reasoning capabilities has further fueled public misconceptions, leading to overestimation of its abilities. It is interesting to find research comparing large language models (LLMs) and human intelligence that does not simply highlight areas where LLMs fall short. Instead, it reveals that LLMs make the same reasoning errors as humans (11). Like people, LLMs struggle with logical reasoning tasks, particularly when faced with abstract stimuli or information that conflicts with prior expectations. This similarity may be due to LLMs learning from human language data, which inherently includes these mistakes. Alternatively, it could indicate that both humans and models develop similar patterns of behavior to enhance predictive accuracy.

There is an issue with how logical ability tests for LLMs are conducted, as they often compare humans and LLMs while assuming that only humans possess true intelligence. This perspective may introduce a bias in assessing the intelligence of LLMs. Do people exhibit bias when evaluating the logical reasoning skills of AI?

**Public Perception of Artificial Intelligence and the formation of stereotype**

Public perceptions of AI are marked by a duality, encompassing both positive and negative views. The ambivalence in public attitudes towards AI reflects a complex emotional landscape regarding its potential impacts, balancing optimism for innovation with caution about its risks and ethical implications. (12-14)

Media narratives significantly influence public attitudes towards AI, leading to a broad but often superficial understanding of the technology. (15, 16) The coexistence of positive and negative portrayals in different regions results in varied public perceptions, highlighting the need for more nuanced and informed discussions about AI. (13)

Without firsthand experience, the public struggles to assess AI's capabilities and limitations comprehensively. Consequently, perceptions are shaped more by external narratives than by rational analysis based on personal experience (16). Perceptions of AI are not fixed; they evolve based on varying levels of trust and understanding. This dynamic can lead to biases and irrational beliefs regarding control over AI technologies (13).

AI's inherent complexity and opacity make it difficult for the public to understand its decision-making processes. Many AI systems are perceived as "black boxes," (17). This lack of transparency reduces trust and acceptance.(18). Uncertainty stemming from data, algorithms, and environmental interactions further affects public confidence in AI (19).Ethical concerns and safety risks associated with AI applications in high-stakes areas like healthcare and justice amplify public caution toward the technology (18).

Technical experts should consider both technological and socio-technical impacts of AI to develop collaborative frameworks (20). As AI continues to evolve and permeate various social domains, ongoing research into public perceptions of AI is necessary (13).

A stereotype is a mental conception that the perceiver develops about a target (21, 22). People tend to rely on stereotypes to judge individuals when they lack information about the target's unique characteristics (23) or when their cognitive processing capacity is limited (24). Stereotypes provide conceptual frameworks that enhance processing efficiency (25-27).

However, stereotypes and prejudice are highly interrelated[4]. Stereotypes are often automatically activated, operating without conscious awareness (28-30). When individuals are unable to consciously monitor stereotype activation, they are more likely to produce stereotype-congruent evaluations of ambiguous behaviors (28).

Given that public perceptions of AI are often superficial, what stereotypes underlie people's views on artificial intelligence? Furthermore, how do these stereotypes influence their judgment of whether AI can truly think?

When individuals do not consciously regulate their stereotypes, the automatic activation of these stereotypes can lead to responses that align with or reflect prejudice, regardless of an individual's explicit level of prejudice. However, controlled inhibition can counteract the influence of automatic stereotype processing, especially when such processing conflicts with the desire to maintain a non-prejudiced self-image (28). Therefore, it is essential to recognize the discrepancies that can arise between individuals' expressed attitudes and their less conscious behaviors. A method originally developed to examine perceptual bias in media (31) is applied here to assess people's beliefs about AI's reasoning capabilities.

**The purpose and scope of the research**

The logical reasoning ability of AI is a key aspect used to assess its intelligence, as demonstrated by the logical arguments presented in AI-generated text. The logical argumantation of a text can be evaluated using objective criteria, regardless of whether the evaluator comprehends the underlying technology of AI. Therefore, it is feasible to examine the perceived biases of individuals with varying viewpoints on generative AI by analyzing their assessments of its logical argumentation capabilities.

This study investigates whether people have bias when they evaluate the logical argumentation ability of AI writing. What factors influencing people's perceptions of AI's logical reasoning capabilities?  Whether their evaluations of AI-generated texts are shaped by preconceived attitudes toward AI technology. To address these research questions, the study employs a two-part design comprising an experiment and a questionnaire.

In the experiment, participants were asked to evaluate the logical reasoning of two texts without knowing whether they were written by a human or generated by AI. Participants were divided into two groups: one received prior instruction on logical reasoning, while the other did not. Each participant read two texts on the same topic—one human-authored and the other AI-generated. In addition to their evaluations, participants answered three open-ended questions: two about the reasons for their evaluations and one about their perceptions of AI writing. Finally, they were asked to identify which text they believed was more likely written by AI. The experiment aims to examine whether participants' differing perspectives lead to different evaluation outcomes and to identify factors influencing judgments of AI-generated content's logical reasoning. The core hypothesis posits that negative preconceptions about AI's reasoning abilities may result in lower ratings for AI-generated texts, even when content quality is comparable. This study tests whether biases affect the objective evaluation of AI writing quality.

---

[4] Bias and prejudice are terms used to describe a set of beliefs that influence and sometimes misinform decisions and interactions with others. Bias is a preference for or against a person, idea, or thing. Prejudice is judging or forming an opinion before having all of the relevant facts.

The second part of the study repeated the evaluation test, with participants again tasked with assessing the logical argumentation of two texts and explaining their ratings. A questionnaire was designed to quantify participants' attitudes toward AI, based on insights from the first part of the experiment. It included 23 questions across five sections, focusing on key attitudinal factors that might influence evaluations of AI-generated texts. This quantitative study treated participants' overall perceptions of AI writing as the independent variable and the rating differences between AI- and human-written texts as the dependent variable, aiming to establish a causal relationship between these attitudes and evaluation outcomes.

## RESULTS

### Part 1: Experimental Results

**Identify AI writing :**
For the group received prior instruction about logical reasoning (group 1), about 93% of the participants correctly recognized that Text 1 was writtern by AI. For the group without prior instruction of logical reasoning (group 2), about 90% of the participants correctly recognized that Text 1 was writtern by AI. This means it is not difficult to identify AI writing.

**Score Analysis:**
For the group received prior instruction about logical reasoning (group 1), Text 1 received an average score of 3.46, while Text 2 scored higher at 3.78. For the group without prior instruction of logical reasoning (group 2), Text 1 received an average score of 3.56, while Text 2 scored higher at 4.20. No significant differences were found in mean scores or variances of Text 1 between the two groups. Group 2 rated Text 2 significantly higher compared to the group 1. But for both groups, Text 2 scored higher than Text 1. But only for group 2 the difference between Text 1 and Text 2 is significant. Statistical analysis compared group 1 and group 2 ratings for the two texts (see Table 1).

Table 1. Statistical analysis of group 1 and group 2's ratings of two texts

|  | T1_group 1 | T1_group 2 | T2_group 1 | T2_group 2 | T1-T2_group 1 | T1-T2_group 2 |
| --- | --- | --- | --- | --- | --- | --- |
| Mean | 3.46 | 3.56 | 3.78 | 4.20 | -0.32 | -0.64 |
| Variance | 0.73 | 1.02 | 1.14 | 0.50 | 2.02 | 1.50 |
| T-test | 0.573 |  | 0.007** |  | 0.050 | 0.001*** |
| F-test | 0.189 |  | 0.002** |  | 0.047 | 0.001*** |

Comparing the scores of the two texts, as shown in Figure 2, more often the score for Text 1 is lower than that for Text 2.

In group 1:
- 25 cases (31%) Text 1 had a higher lower than that for Text 2
- 11 cases (14%) Text 1 had a score equal to that for Text 2
- 44 cases (55%) Text 1 had a score lower than that for Text 2

In group 2
- 8 cases (16%) Text 1 had a higher lower than that for Text 2
- 15 cases (31%) Text 1 had a score equal to that for Text 2
- 26 cases (53%) Text 1 had a score lower than that for Text 2

**Bias analysis**

The analysis of participants' reasoning for rating the two texts reveals a bias in perception. When Text 1 receives a higher rating, its strengths are emphasized, while the weaknesses of Text 2 are amplified. In the first group, 36% of the reasons for evaluating Text 1 are entirely positive, while 60% of the reasons for evaluating Text 2 are entirely negative (see Table 2 and Figure 3). In the second group, 50% of the reasons for Text 1 are entirely positive, and 50% for Text 2 are entirely negative (see Table 3 and Figure 4).

When Text 1 receives a low rating, its disadvantages are magnified, its advantages overlooked, while the advantages of Text 2 are emphasized, and its disadvantages ignored. In the first group, 32% of the reasons for evaluating Text 1 are entirely negative, while 82% of the reasons for evaluating Text 2 are entirely positive (see Table 2 and Figure 3). In the second group, 54% of the reasons for Text 1 are entirely negative, and 69% for Text 2 are entirely positive (see Table 3 and Figure 4).

When the ratings of Text 1 and Text 2 are equal, the reasons for evaluation tend to be more balanced, with both advantages and disadvantages considered. However, some evaluators believe both texts are entirely positive, focusing only on their strengths. In the first group, 73% of the reasons for evaluating Text 1 were critical, compared to 64% for Text 2 (see Table 2 and Figure 3). In the second group, 47% of the reasons for Text 1 were critical, while 20% were critical for Text 2 (see Table 3 and Figure 4).

Therefore, in the first two conditions, evaluators appeared to show greater bias in their ratings of the two texts. In contrast, under the third condition, evaluators were more critical and exhibited less bias in their assessments.

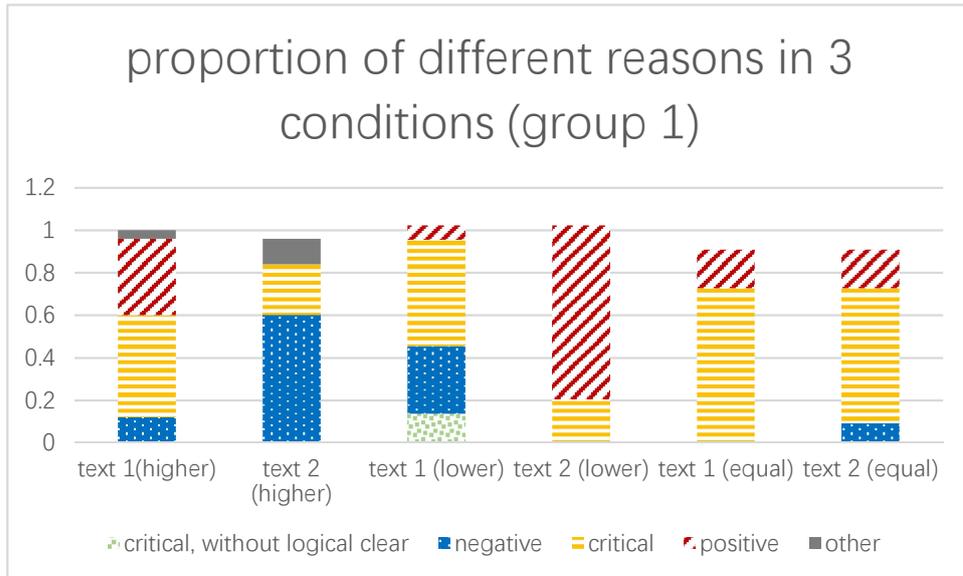

Figure 1. The proportion of different reasons in 3 conditions (group 1)

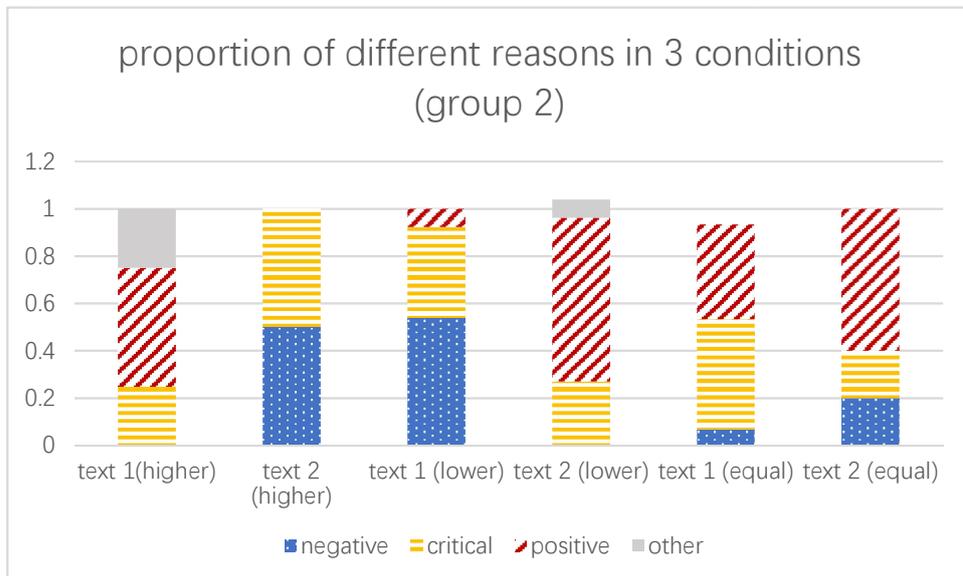

Figure 2. The proprotion of different reasons in 3 conditions (group 2)

Table 2 The proportion of different reasons in 3 conditions (group 1)

|  | text 1(higher) | text 2 (higher) | text 1 (lower) | text 2 (lower) | text 1 (equal) | text 2 (equal) |
| --- | --- | --- | --- | --- | --- | --- |
| critical, without logical clear | 0 | 0 | 0.136364 | 0 | 0 | 0 |
| negative | 0.12 | 0.6 | 0.318182 | 0 | 0 | 0.090909 |
| critical | 0.48 | 0.24 | 0.5 | 0.204545 | 0.727273 | 0.636364 |
| positive | 0.36 | 0 | 0.068182 | 0.818182 | 0.181818 | 0.181818 |
| other | 0.04 | 0.12 | 0 | 0 | 0 | 0 |

Table 3. The proportion of different reasons in 3 conditions (group 2)

|  | text 1(higher) | text 2 (higher) | text 1 (lower) | text 2 (lower) | text 1 (equal) | text 2 (equal) |
| --- | --- | --- | --- | --- | --- | --- |

| negative | 0 | 0.5 | 0.538462 | 0 | 0.066667 | 0.2 |
| --- | --- | --- | --- | --- | --- | --- |
| critical | 0.25 | 0.5 | 0.384615 | 0.269231 | 0.466667 | 0.2 |
| positive | 0.5 | 0 | 0.076923 | 0.692308 | 0.4 | 0.6 |
| other | 0.25 | | 0 | 0.076923 | | |

Training in logical reasoning has a noticeable impact on the evaluation of logical argumentation. Subjects who received training demonstrated more critical thinking and were better at distinguishing between the two passages. Their evaluation reasons were more aligned with the logical argumentation in the texts. In contrast, subjects without training tended to focus more on semantics and personal feelings, often overlooking logical arguments.

**Textual Analysis:**
Semantic analysis divided participants into three conditions based on their scoring preferences:
1. Those rating Text 1 lower than Text 2.
2. Those rating Text 1 higher than Text 2.
3. Those assigning equal scores to both texts.

In the three cases, text analysis was performed on the answers to three open-ended questions: the reasons for the score of Text 1, the reasons for the score of Text 2, and the opinions on AI writing, to find out their commonalities. In the semantic analysis, the content of group 1 and group 2 were combined.

**1) Higher Ratings for Text 1:**
when Text 1 receives a higher score than Text 2, evaluators believe Text 1 excels in logic, depth, evidence support, language expression, and persuasiveness, leading to its higher rating.

**Logic:**
Text 1 is considered to have a clear and coherent logical structure, while the second lacks such clarity and coherence.
Text 1 tightly focuses on the central argument, whereas the second strays from the argumentative focus.
Evaluators believe Text 1 uses concrete evidence and cites relevant studies to support its points, while the second lacks examples and data. (Although Text 1 cites studies in its text, these references are AI-generated fabrications and lack actual sources. Evaluators have overestimated the reliability of the evidence provided by the AI.)
Text 1 has a clear reasoning process and is more persuasive, while the second lacks clarity in its reasoning and persuasive power.

**Depth:**
Text 1 is considered to provide a more comprehensive and systematic analysis, with sufficient explanations and evidence supporting its arguments. In contrast, Text 2 is seen as having a narrower perspective and failing to adequately explain or substantiate certain key points.

**Style:**
Text 1 is viewed as more precise and concise, while the second is perceived as vague and verbose.

**Attitude to AI writing:**
When Text 1 scores higher than the second, evaluators' views on AI writing can be categorized into four aspects: efficiency, logical structure, creativity, and style. Their feedback suggests that while AI writing is recognized for its efficiency and logical clarity, it has limitations in creativity and emotional expression. Evaluators believe AI writing enhances efficiency and speed, with clear logical structures and coherent arguments. However, it falls short in creativity, originality, analytical depth, emotional richness, and personal style. Evaluators emphasized AI's efficiency without expressing significant concerns about dependency. In terms of style, they noted the lack of emotional and personal touch but did not view its concise expression negatively.

**2) Lower Ratings for Text 1:**
Evaluators highlighted Text 2's superior reasoning, depth, structure, and engaging style. Common criticisms of Text 1 included a lack of evidence, reasoning, and appeal. When the score for Text 1 is lower than that of Text 2, evaluators believe Text 2 surpasses Text 1 in terms of logical reasoning, analytical depth, structure, and style, which justifies the higher rating.

**Logic:**
Regarding logical reasoning, evaluators gave opposing comments on the two pieces. They pointed out that Text 1 lacked a reasoning process and specific examples, whereas the second piece excelled in these areas. Additionally, the second piece provided abundant evidence, making it more convincing.

In terms of structure, they felt Text 1 was not cohesive, partly due to the absence of a reasoning process. Conversely, the second piece was seen as more coherent and smooth. (The evaluators associated coherence with logic, albeit misunderstanding that smoothness equates to better logical reasoning.)

**Depth:**
For analytical depth, Text 1 was considered shallow, broad, and lacking detailed discussion. In contrast, the second piece was deemed more profound, offering in-depth and comprehensive analysis from multiple perspectives. This may also reflect their impression that the first piece lacked reasoning process and examples.

**Style:**
On style, Text 1 was perceived as dull, unengaging, abstract, and overly theoretical, failing to connect closely with practical matters. On the other hand, the second piece was more natural, fluent, engaging, and relevant to real-life scenarios, incorporating empirical evidence and thus appearing more convincing. (Evaluators showed a clear preference for certain writing styles.)

**Attitude to AI writing:**
When the first piece is rated lower than the second, commonalities in user evaluations of AI writing can be summarized into five aspects: efficiency, logical reasoning and analytical depth, creativity, . These perspectives indicate that while AI writing is recognized for its efficiency, it has limitations in creativity, logic, and depth. Evaluators believe AI struggles to generate truly innovative and original content and often falls short in logical reasoning and depth of analysis. Concerns about over-reliance on AI are prevalent, with the view that AI writing should serve as a supplementary tool to human creativity.

### 3) Equal Ratings:

When Text 1 receives the same score as Text 2, the common reasons for the ratings suggest that evaluators gave critical assessments of the texts in terms of logic, analytical depth. Key points include:

**Logic:** Text 1 is considered to have a clear and coherent logical structure, but lacked a reasoning process and specific examples. Text 2 provided abundant evidence with clear reasoning process, making it more convincing, but it lacked statistical data and the argumant astray from the center.

**Depth:** Text 1 is considered to provide a more comprehensive and systematic analysis, but lacking detailed discussion. Text 2 was deemed more profound, offering in-depth reasoning analysis, but is seen as having a narrower perspective.

**Style:**

Text 1 is more precise and concise, but lacks emotions. Text 2 is natrual and connect closely with practical matters, but it is lengthy and unrefined.

**Attitude to AI writing:**

When the scores for the two texts are equal, evaluators' opinions on AI writing share the following commonalities in cluding efficiency, logical reasoning, creativity

Overall, these shared perspectives reflect a balanced view of AI writing, recognizing its potential benefits while acknowledging its limitations. Most evaluators support using AI as an auxiliary tool, emphasizing the critical role of humans in the writing process.

In terms of logical reasoning, evaluators find AI writing to be strong in logic but still flawed in handling complex reasoning. On the whole, their views are more positive, resembling those of evaluators who rated the first piece higher than the second.

AI lacks creativity, emotional expression, and the ability to handle complex reasoning and argumentation.

### Findings of Textual Analysis

Through the analysis of the questionnaire responses, differences in ratings were observed: approximately 54% of evaluators considered the second piece of writing (human-written) superior to the first (AI-written), while about 25% believed the first piece was better than the second one. And about 20% have critical views on both texts. indicating that more people still don't believe that artificial intelligence can write like humans..

Evaluators "perceive" the content differently, leading to disagreements about the nature of the stimuli they assessed (Vallone, Ross, and Lepper, 1985). Those who rated human writing higher criticized AI writing for lacking coherent logic, concrete evidence, reasoning processes, and depth. In contrast, evaluators who rated AI writing as comparable or superior highlighted its stronger logical consistency and clearer structure. These opposing views on AI reflect their evaluations and the differences in their ratings of the two texts. Neutral judges, however, considered both the strengths and weaknesses of each article.

However, both groups exhibited misunderstandings. Those who preferred human writing failed to recognize the logical structure in AI writing, dismissing logically sound sections as lacking logic. They also misinterpreted the lack of logical reasoning and the off-topic content in the human-written piece as smooth, coherent, and logical. Their assessment replaced the concept of logical argumentation with reasoning steps and a sense of coherence while reading. Meanwhile, those who considered AI writing not inferior to human writing failed to identify fabricated evidence, mistaking it for genuine research data.

Both groups shared similar views on AI's limitations in innovation, emotional expression, and conveying personal style. However, they differed in their assessments of logic, depth, and writing style. Those who viewed AI writing as comparable or superior had a more positive perception of AI overall.

### Inspiration and limitation of the experiment

The qualitative analysis of open-ended responses suggests that people's evaluations of AI writing's logical reasoning are not based on a clear understanding of the concept of logical argumentation but rather on a vague impression likely influenced by their broader opinions of AI writing, particularly its perceived logical qualities.

The results also highlight the incomplete nature of evaluators' comments on AI writing. Not all responses addressed aspects such as logic, analytical depth, style, creativity, efficiency, or dependency. Among the evaluations regarding logic, 32 gave positive feedback, 15 were negative, and 81 offered no comments. This limitation of open-ended questions shows that evaluators may not consider all aspects comprehensively, even if they hold opinions on the omitted areas.

To address this, further research should build on these qualitative findings to conduct more systematic quantitative studies on AI writing. By providing specific descriptive options for areas such as logic, efficiency, style, creativity, and attitudes, evaluators can give more structured and comprehensive feedback.

## Part 2: Questionaire Results

### The average score for the two texts

The average score for Text 1 was 3.36, while the average score for Text 2 was 3.6, indicating a higher evaluation of Text 2. The score distributions for the two texts are shown in Figure 4. Overall, the data suggests that most evaluators held a higher opinion of Text 2.

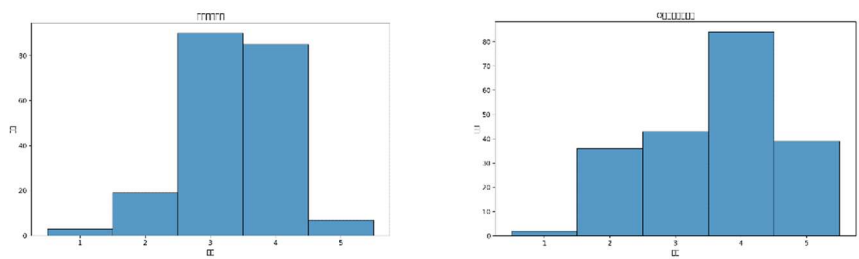

Figure 3: Distribution of scores for Text 1 (left) and Text 2 (right).

### Comparison of Scores for Both Texts

- Cases where the score for Text 1 was lower than the score for Text 2: 102 (50.00%)

- Cases where the scores for both texts were equal: 37 (18.14%)
- Cases where the score for Text 1 was higher than the score for Text 2: 65 (31.86%)

under the three conditions the data was analyzed to compare their means and standard deviations across five dimensions: efficiency, logic, style, creativity, and attitude. As shown in Table 4, the means for all five dimensions in Category 2 were higher than those in Category 1. This indicates that evaluators in Category 2 rated AI writing more favorably in these five aspects.

Table 4: Means and Standard Deviations of Efficiency, Logic, Style, Creativity, and Attitude under three conditions

| Variable | Statistics | T1 higher | T1 equal | T1 lower |
| --- | --- | --- | --- | --- |
| Logic | Mean | 3.46 | 3.78 | 3.76 |
|  | Var | 0.61 | 0.58 | 0.56 |
| Efficiency | Mean | 3.39 | 3.50 | 3.57 |
|  | Var | 0.49 | 0.50 | 0.67 |
| Style | Mean | 2.82 | 2.94 | 2.94 |
|  | Var | 0.48 | 0.36 | 0.56 |
| Attitude | Mean | 3.33 | 3.33 | 3.50 |
|  | Var | 0.64 | 0.64 | 0.69 |
| Creativity | Mean | 3.09 | 3.16 | 3.16 |
|  | Var | 0.53 | 0.49 | 0.62 |

To examine the influence of the five dimensions on the classification of scores, a regression analysis was conducted using the values of logic, efficiency, style, creativity, and attitude as independent variables and the score classification as the dependent variable.

The results, ranked by significance, are shown in Table 5:

Table 5. The influence of the five factors on the score classification

| Factor | coef | std err | t | P>|t| | [0.025 | 0.975] |
| --- | --- | --- | --- | --- | --- | --- |
| Logic | -0.3423 | 0.101 | -3.398 | 0.001 | -0.541 | -0.144 |
| Efficiency | -0.2254 | 0.111 | -2.037 | 0.043 | -0.443 | -0.007 |
| Style | -0.1928 | 0.127 | -1.515 | 0.131 | -0.444 | 0.058 |
| Attitude | -0.1462 | 0.094 | -1.552 | 0.122 | -0.332 | 0.040 |
| Creativity | -0.0912 | 0.113 | -0.805 | 0.422 | -0.314 | 0.132 |

Only the logic factor had the most significant influence on score differences ($P < 0.01$). The influence of other factors is not significant. These findings validate the study hypothesis that evaluators' perceptions of AI writing, particularly regarding its logical reasoning capabilities, significantly influence their ratings of the texts.

**Analysis Based on AI Usage Frequency**

Given that the frequency of AI usage is likely correlated with evaluators' perceptions of AI writing, a comparative and regression analysis was conducted on the data based on usage frequency. Table 6 shows the classification of five factors according to usage frequency.

Table 6. Means and Standard Deviations of the five factors according to usage frequency

| Variable | Statistics | Frequently used | Occasionally used |
|---|---|---|---|
| Efficiency | Mean | 3.58 | 3.27 |
|  | Var | 0.34 | 0.21 |
| Logic | Mean | 3.69 | 3.49 |
|  | Var | 0.37 | 0.33 |
| Style | Mean | 2.89 | 2.86 |
|  | Var | 0.27 | 0.18 |
| Creativity | Mean | 3.15 | 3.08 |
|  | Var | 0.30 | 0.30 |
| Attitude | Mean | 3.48 | 3.20 |
|  | Var | 0.49 | 0.29 |

The difference in efficiency factor is more obvious (3.5758 vs 3.2708), followed by the difference in attitude factor (3.4803 vs 3.2), the difference in logic factor is also obvious (3.6851 vs 3.4861), and the difference in style factor is the smallest (2.8879 vs 2.8639). In terms of variance, the group with higher frequency of use mostly has larger variance, indicating that the data distribution is relatively more dispersed. Only the variance of the creativity factor is very close between the two groups (0.3043 vs 0.3046).

A regression analysis was conducted using the values of logic, efficiency, style, creativity, and attitude as independent variables, and the frequency of usage as the dependent variable. The results, as shown in Table 7, revealed the following:

Table 7. The relationship between perceptions of AI writing and frequency of use

| Variable | Coefficient | Std_Error | Z_Value | P_Value | CI_Lower | CI_Upper | Pseudo_R2 | Log_Likelihood |
|---|---|---|---|---|---|---|---|---|
| Logic | -0.5649 | 0.254 | -2.226 | 0.026* | -1.062 | -0.068 | 0.0195 | -129.865 |
| Efficiency | -1.231 | 0.343 | -3.59 | 0*** | -1.903 | -0.559 | 0.06 | -124.506 |
| Style | -0.1012 | 0.301 | -0.336 | 0.737 | -0.692 | 0.489 | 0.0004 | -132.39 |
| Creativity | -0.2137 | 0.268 | -0.798 | 0.425 | -0.739 | 0.311 | 0.0024 | -132.127 |
| Attitude | -0.691 | 0.243 | -2.845 | 0.004** | -1.167 | -0.215 | 0.0331 | -128.066 |

- Efficiency had the most significant correlation with usage frequency (P < 0.001).
- Attitude followed as the next most significant factor (P = 0.004 < 0.01).
- Logic also showed a significant correlation (P = 0.026 < 0.05).
- Creativity and style factors were not significantly related to usage frequency.

These findings suggest that individuals who frequently use AI tend to have more positive perceptions of its efficiency, attitude, and logical reasoning capabilities, which in turn may influence their overall evaluation of AI writing.

## DISCUSSION

### A biased perceptioin of AI rooted in anthropocentrism

Identifying which text is generated by AI is relatively straightforward, with 90% of individuals making accurate judgments. However, biases in perceptions of AI's logical reasoning capabilities are evident. Those who believe that AI lacks the ability to think— viewing its logical reasoning as inadequate—tend to overrate the logical reasoning skills of human-written texts. Their evaluations are influenced more by comparisons of reasoning styles between AI and humans than by objective criteria. Conversely, individuals who consider AI capable of thinking—believing its logical reasoning to be sound—tend to overestimate AI's reasoning abilities and are more likely to critique flaws in human arguments. Neutral observers, on the other hand, are generally able to identify the strengths and weaknesses of both AI and human reasoning.

It is supposed that subjects find it relatively easy to identify AI-generated text based on writing style. However, their scores on the logical reasoning ability of the text are primarily linked to their opinions about AI's logical reasoning capabilities, rather than the writing style itself. This is indicated by the result that style does not influence their judgment of AI's reasoning ability. Their evaluations are not grounded in objective criteria; instead, they reflect pre-existing beliefs, likely stemming from the notion that machines cannot think. Despite over a century of debate since Ada Lovelace first questioned machine intelligence and Turing proposed a testing method, neither side has provided definitive evidence (2). Consequently, people's beliefs on this matter shape their perceptions of reality, a point we have confirmed in this article.

In the study, a majority of participants rated human-authored texts more favorably, reflecting widespread skepticism about machines' ability to think. One evaluator's open-ended response illustrated this tension: "Oops, the first one was written by AI, but I prefer the first one." This comment suggests that the evaluator believes AI is incapable of logical reasoning or that AI's writing should not match the quality of human writing. However, his actual preference for the AI-generated text creates a state of cognitive dissonance.

Perceptions of AI's logical reasoning capabilities are closely linked to people's beliefs about whether machines can think like humans. For many, "thinking" is synonymous with logical reasoning, as they struggle to conceive of intelligence beyond the framework of human cognition. A basic understanding of AI principles often leads to skepticism about its ability to think like humans, given that AI relies on probabilistic predictions (32). However, few realize that human consciousness also operates on a probabilistic basis (33, 34). This raises an intriguing question: could there be alternative forms of intelligence?

In fact, whether one believes that machines cannot think like human or that AI systems can be not only intelligent, but also conscious, both perspectives are tied to the cognitive bias of anthropocentrism. It is the tendency to place humans at the centre of things and to interpret the world in terms of human values and experiences (34). A closer examination of the definition of "intelligence" reveals that it is not exclusively limited to humans. It is ability to achieve goals in a wide range of environments (34, 35). "Reasoning, goal-directed planning, and linguistic competence are all aspects of intelligence, but non-human animals that lack these competences may still exhibit intelligent behaviour."(34)

Some critics argue that the Turing test fails to address the issue of machine consciousness, specifically whether artificial intelligence possesses any degree of subjective experience or even semantic understanding. One fundamental difference between machine intelligence and human intelligence lies in humans' embodied consciousness, which allows them to provide context and meaning to language through various methods—a capability that is difficult for computers to replicate. Machine intelligence has never possessed embodied consciousness (36, 37).

**The effect of the training of Logical reasoning**

When comparing Group 1 (with pre-training in logical reasoning) to Group 2 (without pre-training), the results suggest that the effect of the training is limited. Evaluators with training were relatively more critical, and more of them were able to identify both the advantages and disadvantages in the two texts. Their evaluation reasons were more aligned with the logical argumentation in the texts. In contrast, those without training tended to focus more on semantics and personal feelings, often overlooking the logical arguments.

However, the training did not change participants' preconceived opinions about whether AI can think. The results showed no significant difference in the ratings for Text 1 between Group 1 and Group 2, nor in the comparison between the two texts. The primary difference was observed in ratings for Text 2 (the human-authored text), where Group 2 assigned higher scores. Group 2 also provided more positive reasons for Text 2 without identifying its logical reasoning weaknesses. Overall, the evaluators did not base their evaluations on objective criteria.

**Usage frequency influences people's attitudes toward AI**

While concerns exist that excessive reliance on AI may weaken independent thinking, this study found that frequent users of AI tend to hold more positive attitudes toward it. Individuals who frequently use AI tend to have more positive perceptions of its efficiency. An increase in frequency of use improves users' attitudes toward AI and strengthens their willingness to integrate AI into their work.They are more willing to integrate AI into writing tasks and believe that using AI does not impair their ability to think independently. Instead, the process of guiding AI to generate content that aligns with their expectations requires continuous critical thinking, helping them maintain cognitive engagement (Alice Barana et al., 2023). In contrast, those who limit AI usage out of concern may struggle to validate their perceptions through hands-on experience, solidifying negative impressions.

Generative AI powered by large language models is not a traditional tool but a system that can continually learn and improve through user feedback. In its early stages, AI capabilities may be limited, but user guidance and adjustments help it evolve to better meet user needs. This process not only aids in AI improvement but also enhances user trust, thereby promoting wider adoption. This confidence-boosting cycle is not only applicable to writing but extends to other domains of AI application as well.

Attitudes significantly influence how people engage with AI. Those who approach AI with skepticism are more likely to focus on negative aspects that confirm their biases, such as mocking early instances of nonsensical ChatGPT responses. This reinforces distrust and reduces usage, potentially hindering cognitive acceptance of technological advancements. Conversely, those with a curious and exploratory mindset are more inclined to test AI's performance in various tasks. Even if initial results are unsatisfactory, curiosity drives them to persist, enabling a deeper understanding of AI's strengths and weaknesses and fostering receptiveness to technological breakthroughs.

# MATERIALS AND METHODS

## Part 1: Experiment on the evaluation of logical argumentation ability of AI writing

**Participants and Sample:**

The experiment involved a convenience sample of 129 students, drawn from four second-year university writing classes and two graduate-level courses taught by the researchers. While not randomized, this sample offered the advantage of consistency, as participants shared similar levels of text analysis skills, reducing variability in the results.

Table 8. Basic information of the subjects

| Class | Number | Educational level | Course |
|---|---|---|---|
| Class 1 | 18 | Bachelor's Degree | Writing and Communication |
| Class 2 | 18 | Bachelor's Degree | Writing and Communication |
| Class 3 | 22 | Bachelor's Degree | Writing and Communication |
| Class 4 | 22 | Bachelor's Degree | Writing and Communication |
| Class 5 | 22 | Master's Degree | Introduction to Creative Thinking |
| Class 6 | 27 | Master's Degree | Introduction to Creative Thinking |
| **Total** | **129** | | |

**Experimental Materials:**

The experiment involves a questionnaire that asks participants to evaluate two texts of comparable length on the same topic—analyzing the lack of critical thinking skills among Chinese children from the perspective of family education. The two texts are from two versions of the paper submitted by the same student under the same topic. The first version was written using AI, while the second version was written by the student himself. The full paper is around 5,000 words in Chinese, and only a section is excerpted here, focusing on showcasing two different logical argumentation styles. The first text, written by Wenxin Yiyan—an AI developed by Baidu, a leading Chinese technology company—is characterized by a clear structure, precise reasoning, and logical coherence, but it lacks specific evidence, reasoning processes, and engaging language, and contains fabricated evidence. The second text, written by a human, is more fluent and vivid, includes familiar examples, and has emotional appeal but is overly opinionated and lacks effective logical reasoning. This type of experimental material is used to test whether people can accurately perceive these mixed features or whether they ignore some of them because they only pay attention to the expected features.

The questionnaire comprises eight questions:
1. Basic information about participants (Q1-Q2).
2. Ratings for the first text (Q3, on a 1-5 scale) and reasons for the rating (Q4).
3. Ratings for the second text (Q5, on a 1-5 scale) and reasons for the rating (Q6).
4. Participants' perceptions of AI writing (Q7).
5. Identification of which text is more likely written by AI (Q8).

**Experimental Procedure:**

The experiment was conducted in class. For the first four writing classes, the author provided a 30-minute lecture on logical argumentation before the students evaluated the texts. The remaining two non-writing classes evaluated the texts without prior instruction. Students in Classes 1 and 2 completed paper-based questionnaires, while those in Classes 3-6 used electronic surveys for easier data handling. All questionnaires were completed within 30 minutes under the researcher's supervision.

Responses from paper-based questionnaires were manually input into spreadsheets, while data from electronic questionnaires were downloaded directly. Collected data included both quantitative scores and qualitative answers to open-ended questions.

**Data analysis:**

Data analysis employed the AI tool "Julius," which integrates Python for computation via a ChatGPT interface (https://julius.ai/chat?id=f59f1976-5722-48be-9890-de91a2ccf606). We used the stand version of Julius. We use this tool because it is efficient for analyzing qualitative data, extracting keywords, and performing semantic analysis, such as find out the common points in certain conditions. At the same time, the content was examined and classified according to the semantics manually as verification of the AI analysis.

**Part Two: A Quantitative Study on How Perceptions of AI Writing Influence Evaluations of Its Logical Argumentation**

**Participants and Sample:**

This study recruited participants on a voluntary basis. Recruitment information was disseminated via social networking platforms at a Chinese university. A total of 233 participants, all enrolled students at the university, responded voluntarily. Among them:

- **Gender**: 143 male participants (61.37%), 90 female participants (38.63%).
- **Education Level**: 170 undergraduates (72.96%), 58 master's students (24.89%), and 5 doctoral students (2.15%).
- **AI Usage**: 151 participants (64.81%) frequet use generative AI to assist in learning and work, 82 (35.19%) use it occasionally, and 0 reported never using it. All participants had prior experience with AI tools.

Their academic disciplines are shown in Figure 3.

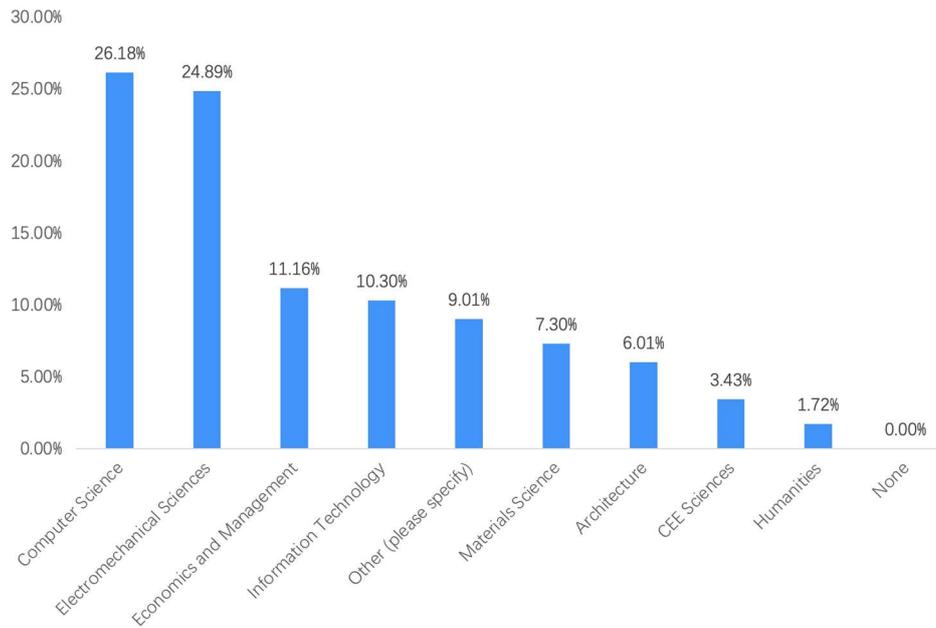

Figure 4. The academic disciplines of the participants

**Research Instruments:**

A custom-made questionnaire was used, similar to the experimental content from the first part of this study. In addition to collecting basic demographic information, the main difference was that the open-ended question about perceptions of AI writing was replaced with 23 closed-ended items. A 6-point Likert scale was employed, ranging from 1 (strongly disagree) to 6 (strongly agree).

Table 9. 23 items on perceptions of AI writing

| Factor | No. | Item |
|---|---|---|
| **Efficiency** | 1 | AI writing enhances efficiency and convenience, facilitating rapid content and idea generation. |
| | 2 | Over-reliance on AI writing may reduce independent thinking and critical analysis skills. |
| **Logic** | 1 | AI writing often lacks depth in logical reasoning and analysis. |
| | 2 | AI writing is frequently perceived as having a clear logical structure. |
| | 3 | AI writing does not follow a reasoning process. |
| | 4 | AI writing demonstrates a clear logical flow from premises to conclusions. |
| | 5 | AI writing lacks specific examples to support logical proofs. |
| | 6 | AI writing excels at listing points and presenting arguments from multiple perspectives. |
| | 7 | AI writing provides clarity in argument structure. |
| **Style** | 1 | AI writing is often rigid and formulaic. |
| | 2 | AI writing lacks the emotional depth found in human writing. |
| | 3 | AI writing does not reflect a personal style. |
| | 4 | AI writing is generally more rational in tone. |
| | 5 | AI writing employs rigorous and precise language. |
| | 1 | AI writing is repetitive and lacks originality. |
| | 2 | AI writing struggles to generate truly creative content. |
| **Innovation** | 3 | AI can assist users in developing better ideas. |
| | 4 | There is optimism about AI's future potential to advance in creativity, depth, and writing quality. |
| | 1 | AI writing cannot fully replace human writing. |
| | 2 | AI is limited to handling simple writing tasks. |
| **Attitude** | 3 | AI writing has largely replaced human writing in some contexts. |
| | 4 | I intend to delegate more writing tasks to AI in the future. |
| | 5 | I currently find AI writing to be unhelpful. |

**Research Procedure:**

The experimenter of this study was not aware of the purpose of the experiment. Participants completed the electronic questionnaire by scanning a QR code at an agreed-upon time and location under the supervision of the experimenter. Participants were required to complete the questionnaire independently and received a small monetary reward upon completion.

**Data analysis:**

Questionnaires submitted in less than 240 seconds were deemed invalid and excluded from the analysis. After filtering, 222 valid questionnaires remained.

Additionally, participants who identified the second text as AI-written in the final question were excluded, leaving 204 valid responses for analysis.

Generative AI analysis tools, e.g., Julius (Standard version) and ChatGPT-4 were used to process the valid responses. with SPSS and Excel employed for validation. We performed regression analyses on the data, calculated the distribution of values, the mean, and the standard deviation, and analyzed the differences between the data points.

**Preregistration and ethics approval**

The study was preregistered at AsPredicted.org (ID 136723); a copy of the preregistration is included in section S9. The study was approved by the ethics boards at the UCL School of Management (ID UCLSOM- 2023- 002) and the University of Exeter (ID 1642263). In formed consent was obtained for both the writer study and the evalu ator study.